\RequirePackage{fix-cm}
\documentclass[smallextended]{svjour3}       
\pdfoutput=1
\smartqed  
\usepackage{graphicx} 
\usepackage{multirow}
\usepackage{amsmath}
\usepackage{float}
\usepackage{subfig}

\usepackage{url}
\usepackage{algorithm}
\usepackage{algorithmic}
\usepackage{verbatim}
\graphicspath{{./figures/}}

\usepackage{color}
\usepackage{amssymb}

\usepackage{marvosym}
\newcommand{\envelope}{(\raisebox{-.5pt}{\scalebox{1.45}{\Letter}}\kern-1.7pt)}

\usepackage[sort&compress]{natbib}
\journalname{Advances in Data Analysis and Classification}

\begin{document}

\title{Time Series Classification by Class-Specific Mahalanobis Distance Measures}

\author{Zolt\'an Prekopcs\'ak \and Daniel Lemire}

\institute{Zolt\'an Prekopcs\'ak \envelope \at Budapest University of Technology and Economics, Hungary \\
              Magyar tud\'osok k\"or\'utja 2, Budapest, H-1117 Hungary \\
              \email{prekopcsak@tmit.bme.hu}, Phone: +36 1 4633119, Fax: +36 1 4633107
           \and
           Daniel Lemire \at LICEF, Universit\'e du Qu\'ebec \`a Montr\'eal (UQAM) \\		
              100 Sherbrooke West, Montreal, QC, H2X 3P2 Canada             
}

\date{Received: date / Accepted: date}

\maketitle
\begin{abstract}
To classify time series by nearest neighbors, we need to specify or learn one
or several distance measures.
 We consider  variations of the Mahalanobis distance measures
 which rely on the inverse covariance matrix of the data.
 Unfortunately --- for time series data --- the covariance
 matrix has often low rank. To alleviate this problem we
 can either use a pseudoinverse, covariance shrinking or
 limit the matrix to its diagonal. We review
 these  alternatives and benchmark them against 
 competitive methods such as the related Large Margin Nearest Neighbor
 Classification (LMNN) and the Dynamic Time Warping (DTW) distance. 
 As we expected, we find that the DTW is superior,  but the
 Mahalanobis distance measures are one to two orders of magnitude faster. 
 To get best results with   Mahalanobis distance measures, we recommend learning
  one distance measure per class using 
 either covariance shrinking or the diagonal approach.
 
 
 
\end{abstract}

\section{Introduction}

Time series are sequences of values measured over time. Examples include
financial data, such as stock prices,  or medical data, such as blood sugar
levels. Classifying time series is an important class of problems
which is applicable to music classification~\citep{springerlink:10.1007/s11634-007-0016-x},
medical diagnostic~\citep{Sternickel2002109} or bioinformatics~\citep{Legrand2008215}.

Nearest Neighbor (NN) methods classify time series efficiently and accurately~\citep{1454226}.
In the 1-NN method, 
we let the unclassified instance be in the same class as its nearest classified neighbor.

We need to specify a distance measure: the Euclidean and Dynamic Time Warping~\citep{1163055} distances are popular choices.
However, we can also learn a distance measure based on some training data~\citep{yang2006distance,weinberger2009distance}.
Given the training data set made of classes of time series instances,
we can either learn a single (global) distance measure, or learn one distance
measure per class~\citep{csatari2010,paredes2000class,1221358}. That is, to compute the distance between a test element and an instance of class~$j$, we use
a distance measure specific to class~$j$. 

Consider a family of time series $\textbf{x}^{(1)}, \textbf{x}^{(2)}, \ldots,
\textbf{x}^{(N)}$ of lengths $n$. 
Because the Euclidean distance is popular for NN classification, it is tempting to consider 
generalized ellipsoid distance measures~\citep{ishikawa1998mindreader}, that is, distance measures of the form
\begin{eqnarray*}
D_M(\mathbf{x},\mathbf{y}) = (\mathbf{x}-\mathbf{y})^{\top} M (\mathbf{x}-\mathbf{y})
\end{eqnarray*}
where $M$ is a positive semi-definite matrix
and $\mathbf{x}, \mathbf{y}$ are two time series of lengths $n$.
When the matrix $M$ is the identity matrix, we recover the (squared) Euclidean distance.
We get the \emph{Mahalanobis distance measure} when  
 we use  the matrix $M$ minimizing the sum of distances between the time series in $S$:
$\sum_{\mathbf{x},\mathbf{y} \in S}D_M(\mathbf{x},\mathbf{y})$ (see~\S~\ref{sec:mathstuff}).
Unfortunately, in the context of time series, solving for such an optimal matrix
often involves inverting a low-rank matrix.

Our main contribution is to survey and compare 
techniques to solve
this mathematical difficulty:
\begin{itemize}
\item We may  require $M$ to be a diagonal matrix.
\item We can use a pseudoinverse.
\item We can apply covariance shrinkage.
\end{itemize}
Moreover, we can either learn one such distance measure for the entire data set, or one distance measure per class.
To our knowledge, there was no attempt  to compare these alternatives in the context of time series. 
After comparing these alternatives, we present two main findings:
\begin{itemize}
\item  
We get significantly poorer classification accuracy  when using pseudoinverses.
Indeed, the pseudoinverse approach generates twice
the error rate of 
the covariance shrinkage
or diagonal-matrix approach. 
\item 
We find that the class-specific Mahalanobis distance measures
are preferable to the global Mahalanobis distance measure. That is, it is best to learn one distance measure per class instead of learning one overall distance measure.
\end{itemize}
We also compare our results with other well established techniques such as Large Margin Nearest Neighbor
 Classification (LMNN) and the Dynamic Time Warping (DTW) distance. We find that even though the DTW has superior classification accuracy, it is one to two orders of magnitude slower than Mahalanobis distance measures. 

\section{Related Works}
  
Consider two time series  $\mathbf{x}$ and $\mathbf{y}$ of lengths $n$. 
 The $i^{\mathrm{th}}$ data point of time series $\mathbf{x}$ is written $x_i$. 
Two of the most common distances
between time series are the Manhattan and Euclidean distances. They are special cases ($p=1$ and $p=2$) of the Minkowski distance: $\sqrt[p]{\sum_{i=1}^n |x_i - y_i|^p}$.
Several other distance measures are used for time series classification. 
\cite{1454226} presented an extensive comparison of these distance measures and
concluded that DTW is among the best measures and that the accuracy of the Euclidean distance
converges to DTW as the size of the training set increases.

In a general Machine Learning setting, \cite{paredes2000class,1221358} compared Euclidean distance with the conventional and class-specific Mahalanobis distance measures. 
 One of our contribution is to validate these generic results on time series: instead of tens of features, we have hundreds or even thousands of values which makes the problem mathematically more challenging: the ranks of our covariance matrices are
 often tiny compared to their sizes. 

More generally, distance metric learning has an extensive literature~\citep{255644,hastie1996discriminant,1598381,short1980new}.
We refer the reader to \cite{weinberger2009distance} for a review.

There are many extensions and alternatives to NN classification. For example,
\cite{1558510} use instance weights to improve classification.
Meanwhile, \cite{1553530} learn a distance measure per instance. More generally, the problem of classifying time series has a long history in statistics~\citep{RH19821, AHG:AHG2137,hastie1996discriminant}.

\subsection{Dynamic Time Warping Distance}

 The Dynamic Time Warping distance (DTW) is a generalization of the Minkowski distance which allows the
 data to be realigned~\citep{itakura1975mpr,sakoe1978dpa}.
To compute the DTW between $\mathbf{x}$ and $\mathbf{y}$, you must find a 
many-to-many matching between the data points in  $\mathbf{x}$ and the data points
in $\mathbf{y}$. That is each data point from one series must be matched with at least
one data point with the other series. One such matching is the trivial one, which maps the
first data point from $\mathbf{x}$ to the first data point in $\mathbf{y}$, 
the second data point in $\mathbf{x}$ to the second data point in $\mathbf{y}$, and so on.
A matching can be written as a list of pairs of indexes with one index in the first time series and one index in the other. For example,  the trivial matching
is just $\Gamma=\{(1,1), (2,2), \ldots, (n,n) \}$.
The Minkowski distance corresponding to a matching is defined as
$\sqrt[p]{\sum_{(i,j)\in \Gamma} |x_i - y_j|^p}$.
Typically, $p$ is either 1 or 2: for our purposes we
choose $p=2$.

For a given $p$, the DTW is defined as the minimal Minkowski distance
over all allowed matchings $\Gamma$. That is
$\mathrm{DTW}(\mathbf{x},\mathbf{y})= \min_{\Gamma}\sqrt[p]{\sum_{(i,j)\in \Gamma} |x_i - y_j|^p}$. We can solve for $\Gamma$ using dynamic programming.
It is required for matchings to be monotonic: 
if both $(i,j)$ and $(i+1,j')$ are in $\Gamma$ then $j'\geq j$,
that is, we cannot warp back in time: if the first index increases, the second index cannot decrease. Because of monotonicity, the DTW is not invariant under permutation of the coordinates.
The DTW between $(0,1,0,2)$ and $(0,1,1,2)$ is one with $\Gamma_1=\{(1,1), (2,2), (3,3), (4,4) \}$ whereas the DTW between $(0,0,1,2)$ and $(0,1,1,2)$ is zero with $\Gamma_2=\{(1,1), (2,1), (3,2), (3,3), (4,4) \}$.
Yet they only differ by the permutation of the second and third data points. 

Unlike many other distance measures, such as the Euclidean distance, the DTW can handle sequences of different lengths.
However, according to
 \cite{ratanamahatana2005tmd}  ``comparing sequences of different lengths and reinterpolating them to equal length produce no statistically significant difference in accuracy or precision/recall.'' In other words, when comparing time series having different lengths,  we may  linearly interpolate them to have the same length without loss of classification accuracy.

As an extension, some matches might be forbidden if
the data points are too far apart~\citep{itakura1975mpr,sakoe1978dpa}.
\cite{Yu20112787} has proposed learning this warping constraint
from the data.
\cite{Gaudin:2006:ATW:1193211.1193725} proposed a weighted 
version of the DTW called Adaptable Time Warping. Instead of
computing $\sum_{(i,j)\in \Gamma} |x_i - y_j|^p$, it computes
$\sum_{(i,j)\in \Gamma} M_{i,j}|x_i - y_j|^p$ where $M$ is some matrix.
Unfortunately, finding the optimal matrix $M$ can be a challenge.
\cite{Jeong2010} investigated another form of weighted DTW
where you seek to minimize the cost
$\sqrt[p]{\sum_{(i,j)\in \Gamma} w_{|i-j|}|x_i - y_j|^p}$
where $\mathbf{w}$ is some weight vector.
Many other variations on the DTW   have been
proposed, e.g.,~\cite{springerlink:10.1007/s11634-006-0004-6}.

One disadvantage is that the DTW fails to satisfy the triangle inequality
($\mathrm{DTW}(\mathbf{x},\mathbf{y})+\mathrm{DTW}(\mathbf{y},\mathbf{z}) \geq
\mathrm{DTW}(\mathbf{x},\mathbf{z})$), hence the DTW is not a metric~\citep{Lemire:2009:FRT:1542560.1542887}.

\subsection{Large Margin Nearest Neighbor (LMNN)}

A conventional distance-learning approach is to find an optimal generalized ellipsoid
distance measure with respect to a specific loss function. 
The LMNN algorithm proposed by \cite{weinberger2009distance} takes a different approach. It seeks to force nearest neighbors to 
belong to the same class and it separates instances from different classes by a large margin. 
LMNN can be
formulated as a semi-definite programming problem~\citep{vandenberghe1996semidefinite}.

Specifically, we begin with a generalized ellipsoid distance measure $D_M(\mathbf{x},\mathbf{y})=(\mathbf{x}-\mathbf{y})^{\top} M (\mathbf{x}-\mathbf{y})$. We must solve for the matrix $M$
given some data set of classified time series $\mathbf{x}^{(1)}, \mathbf{x}^{(2)},\ldots, \mathbf{x}^{(N)}$.
We require $M$ to be positive semi-definite, so that the
distance measure $D_M$ is a pseudo-metric:  it is symmetric, non-negative and it satisfies the triangle inequality ($\sqrt{D_M(\mathbf{x},\mathbf{y})}+\sqrt{D_M(\mathbf{y},\mathbf{z})}\geq \sqrt{D_M(\mathbf{x},\mathbf{z})}$).

Prior to computing $M$, we create two $N\times N$ matrices $y$ and $\eta$. We set $y_{ij}$ equal to one whenever
$\mathbf{x}^{(i)}$ and
$\mathbf{x}^{(j)}$ are in the same class,
otherwise  $y_{ij}=0$.
For all time series $\mathbf{x}^{(i)}$, we find $k$~nearest neighbors under the Euclidean distance that are in the same class. (For 1-NN classification, we set $k=1$.) Whenever
$\mathbf{x}^{(j)}$ is a nearest neighbor of $\mathbf{x}^{(i)}$, we set $\eta_{ij}=1$, otherwise $\eta_{ij}=0$. Both matrices are computed once.

\cite{weinberger2009distance} find the positive semi-definite matrix $M$ by minimizing \begin{eqnarray*}
\sum_{ij} \eta_{ij} D_M(\mathbf{x}^{(i)},\mathbf{x}^{(j)})
+c \sum_{ijl} \eta_{ij}(1-y_{il}) \varepsilon_{ijl}
\end{eqnarray*}
where the sums are over the range of indices
$\{1,2,\ldots, N\}$,
subject to the  constraints
that the $\varepsilon_{ijl}$'s are non-negative and
that \begin{eqnarray*}
D_M(\mathbf{x}^{(i)},\mathbf{x}^{(l)})-D_M(\mathbf{x}^{(i)},\mathbf{x}^{(j)})\geq 1-\varepsilon_{ijl}. 
\end{eqnarray*}
The fixed parameter $c$ is set by cross validation.
\cite{weinberger2009distance} called the variables $\varepsilon_{ijl}$ \emph{slacked variables}: they must be determined along with the matrix $M$. 
Though this problem can be solved using 
a generic solver, 
\cite{weinberger2009distance} found that they could get substantially better speed with a custom solver: we use their software in \S~\ref{sec:experiments}.

\section{Mahalanobis distance measures}
\label{sec:mathstuff}

Given a time series $\mathbf{x}^{(k)}$, we write its 
$i^{\mathrm{th}}$~data point as $x^{(k)}_i$.
We compute the (sample) covariance matrix $C = (c_{ij})$ of a family of time series $\mathbf{x}^{(1)}, \mathbf{x}^{(2)}, \ldots, \mathbf{x}^{(N)}$ of lengths $n$ by $c_{ij} = \frac{1}{N-1} \sum_{k=1}^N (x_{i}^{(k)} -  \bar{x_{i}} )(x_{j}^{(k)} -  \bar{x_{j}} )$ where $N$ is the number of instances and where $\bar x_i$ is the average of the $i^{\mathrm{th}}$~data point of the time series ($\bar x_i=\frac{1}{N}\sum_{k=1}^N x_{i}^{(k)} $).

The Mahalanobis distance measure~\citep{mahalanobis1936generalized} is a special case of the generalized ellipsoid distance measure $D_M(\mathbf{x},\mathbf{y})=(\mathbf{x}-\mathbf{y})^{\top} M (\mathbf{x}-\mathbf{y})$ where $M$ is proportional to the inverse of the covariance matrix $M \propto C^{-1}$. Though the  Mahalanobis distance measure is often defined by setting $M$ to the inverse of the covariance matrix ($M=C^{-1}$), we find it convenient to normalize it when possible so that the determinant of the matrix $M$ is one: $M= (\det(C))^{\frac{1}{n}} C^{-1}$ where $n$ is the length of the time series. 
The Mahalanobis distance measure minimizes the sum of distances between time series
$\sum_{\mathbf{x},\mathbf{y}}D_M(\mathbf{x},\mathbf{y})$ subject  to a regularization constraint on the determinant ($\det(M) =
1$).  In this sense, it is optimal.

When the covariance is non-singular ($\det(C)\neq 0$) then the covariance is positive definite, and so is the matrix $M$: it follows that 
the square root of the generalized ellipsoid distance measure is a metric. That is, we have $D_M(\mathbf{x},\mathbf{y})=0 \Leftrightarrow \mathbf{x}=\mathbf{y}$,  it is symmetric, non-negative and it satisfies the triangle inequality ($\sqrt{D_M(\mathbf{x},\mathbf{y})}+\sqrt{D_M(\mathbf{y},\mathbf{z})}\geq \sqrt{D_M(\mathbf{x},\mathbf{z})}$).
Unfortunately, the
 covariance matrix fails to be  invertible when the number of instances ($N$) is smaller than the number of data points ($n$).
In \S~\ref{sec:computing}, we review some other solutions to address this problem.

\section{Computing Mahalanobis distance measures for time series  classification}
\label{sec:computing}

The covariance matrix may be  singular when the number of instances ($N$) is smaller or about the same as the number of data points ($n$) in the time series. 
 This is a common problem with time series: whereas individual time series might have thousands of data points, there may only be a few labeled time series in each class.

\subsection{Diagonal Mahalanobis distance measures}

The most straight-forward solution is to limit the covariance matrix $C$ to its diagonal~---~thus producing a weighted Euclidean distance measure.
Indeed, 
if we
require that the matrix $M$ be zero outside the diagonal, then restricting the covariance $C$ to its diagonal (that is, setting $M^{-1}\propto \mathrm{diag} ( C )$) minimizes the 
sum of distances between time series.
 As long as the variance of
each data point in our training sets is different from zero~---~a condition
satisfied in practice in our experiments, the problem
is well posed and the result is a positive-definite matrix.
 Hence, the generalized ellipsoid distance measure $D_M(\mathbf{x},\mathbf{y})=(\mathbf{x}-\mathbf{y})^{\top} M (\mathbf{x}-\mathbf{y})$
is a metric.
We normalize $M$ so that its determinant is one.
In such a diagonal case, the number of parameters
to learn grows only linearly with the number of data points in the time series.
In contrast, the number of elements in the full covariance matrix grows 
quadratically. One consequence is that 
the diagonal version of the Mahalanobis distance measure is computed much faster ($O(n)$ vs. $O(n^2)$).

Our version of the  diagonal Mahalanobis distance measure is closely related
to the \emph{standardized Euclidean distance} defined
as the Euclidean distance between the components divided by their standard deviation: the square of the  standardized Euclidean distance between $\textbf{x}$ and $\textbf{y}$ is $\sum_{i=1}^n ((x_i-y_i)/\sigma_i)^2$ where $\sigma_i$ is the standard deviation of the $i^{\mathrm{th}}$~component. However we must multiply the square of the standardized Euclidean distance by the Geometric mean of the variances ($\sqrt[n]{\prod_{i=1}^n \sigma_i^2}$) to get our diagonal Mahalanobis distance measure. This normalization is a consequence of our requirement that the determinant of the matrix $M$ be one: $\det(M)=1$.  It is significant because we may simultaneously use several  distance measures in the class-specific NN classification.

\begin{figure}
\centering

\subfloat[Sample of time series]{\includegraphics[width=0.69\columnwidth]{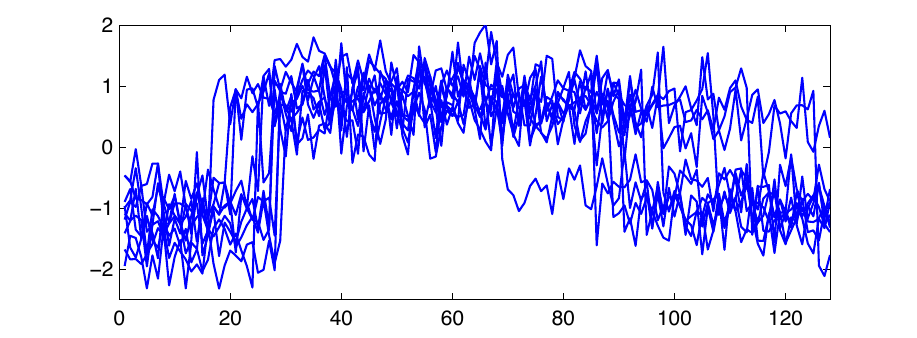}
	}
\subfloat[Sample covariance]{\includegraphics[width=0.29\columnwidth]{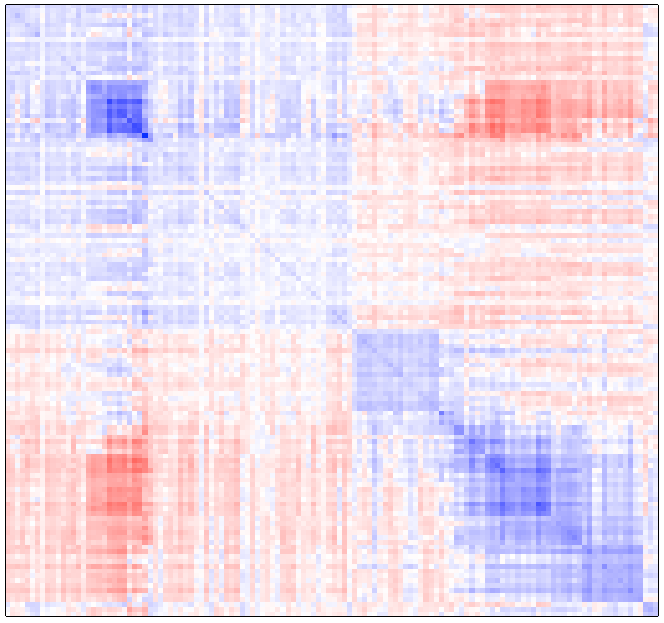}
	}
\caption{\label{fig:signcov} Ten samples of the Cylinder class from the CBF data
set~\citep{921732} and its sample covariance. Each time series has 128~data points.
Higher absolute values in the matrix are presented using darker colors.
}
\end{figure}

 Unfortunately, the diagonal Mahalanobis  distance measure fails to use the information
 off the diagonal in the covariance matrix.
See Fig.~\ref{fig:signcov} for the covariance matrix of a class of time series. It is clear from the figure that the covariance matrix has significant values off the diagonal.
There are even block-like patterns in the matrix corresponding to
specific time intervals.
 
\subsection{Moore-Penrose pseudoinverse and covariance shrinkage}

\begin{figure}
\centering

\subfloat[Sample covariance]{\includegraphics[width=0.3\columnwidth]{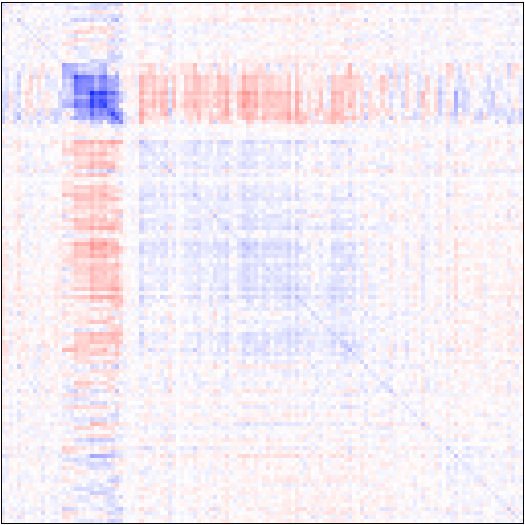}
	}
\subfloat[Shrinked covariance]{\includegraphics[width=0.3\columnwidth]{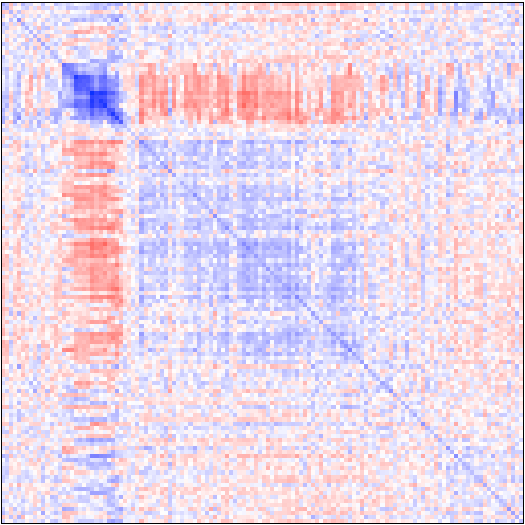}
	}
\subfloat[Diagonal covariance]{\includegraphics[width=0.3\columnwidth]{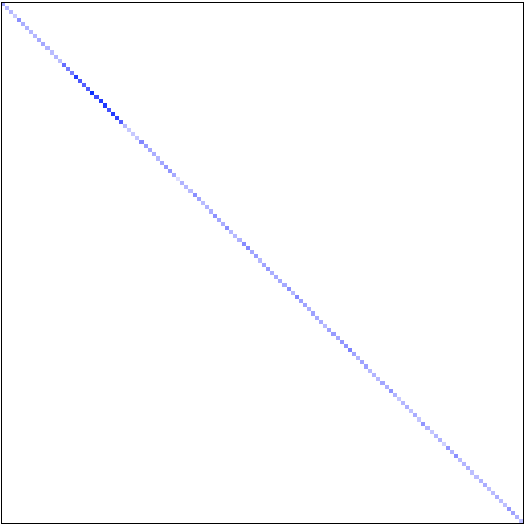}
	}
\caption{\label{fig:signmatrices}
The covariance estimates of the Funnel class in the CBF data set.
Large absolute values are in darker colors.
Both the shrinked
and diagonal covariances are positive definite whereas
the sample covariance matrix is singular. }
\end{figure}

 Could it be that non-diagonal Mahalanobis  distance measures could be superior or at least
 competitive with the diagonal Mahalanobis  distance? 
 It is tempting to use banded matrices, but the restriction of a positive definite
matrix to a band may fail to be positive definite. Block-diagonal matrices~\citep{Matton20101303} can preserve positive definiteness, but learning which blocks
to use in the context of time series might be difficult. 
Instead, we propose two approaches: one is based on the widely used Moore-Penrose pseudoinverse, and the other is covariance shrinkage. 
See Figure~\ref{fig:signmatrices} for the three different covariance estimates of the same class: sample covariance, shrinked covariance and diagonal covariance.

The approach based on the pseudoinverse is based on the singular value decomposition (SVD). We write the SVD as $C= U\Sigma V^{\top}$ where $\Sigma$ is a diagonal matrix with eigenvalues $\gamma_1, \gamma_2,\ldots$ and $U$ and $V$ are orthogonal matrices.
The Moore-Penrose pseudoinverse is given by $V\Sigma^+ U^{\top}$ where $\Sigma^+$
is the diagonal matrix made of the eigenvalues $1/\gamma_1, 1/\gamma_2,\ldots$ with
the convention that $1/0=0$. The pseudo-determinant is the product of the non-zero eigenvalues of $\Sigma$.
We set $M$ equal to the
pseudoinverse of the covariance matrix~---~normalized so that it has a pseudo-determinant
of one. This solution is equivalent to projecting the time series data on the 
subspace corresponding to the non-zero eigenvalues of $\Sigma$. That is, the matrix $M$ is singular.
Since the matrix $M$ is still a positive semi-definite matrix, the square root of the generalized ellipsoid distance measure remains a pseudometric:
it is symmetric, non-negative and it satisfies the triangle inequality. But it is no longer a metric since it is possible to find distinct $\mathbf{x}, \mathbf{y}$ such that $D_M(\mathbf{x},\mathbf{y})=0$.

Covariance shrinkage is an estimation method for problems with small number of instances and large number of attributes~\citep{stein}. It has better theoretical and practical properties for such data sets as the estimated covariance matrix is guaranteed to be non-singular.
The covariance matrix $C$ is positive semi-definite but can be singular. 
To prevent $C$ from being singular, we  replace it
with  an estimation of the form \begin{eqnarray*}C^{\star}=\lambda T + (1- \lambda) C\end{eqnarray*} for some suitably chosen
target matrix $T$: if $T$ is a  positive definite matrix and $\lambda\in (0,1]$, we have that  $\lambda T + (1- \lambda) C$ 
must be positive definite. Moreover, 
the smallest eigenvalue of $\lambda T + (1- \lambda) C$ must be at least as large as $\lambda$ times the smallest eigenvalue of $T$. 
We have used the target recommended by \cite{schafer2005shrinkage} which is the diagonal of the unrestricted covariance estimate, $T = \mathrm{diag} ( C )$.
It is positive definite in our examples. For $\lambda$, we use the estimation proposed by \cite{schafer2005shrinkage} (see Appendix~\ref{sec:choiceoflambda} for details).
We then set $M^{-1} \propto C^{\star}$, normalizing so that $\det (M)=1$.
Unlike the pseudoinverse approach, covariance shrinkage generates a generalized ellipsoid distance measure which is a metric.

\subsection{Global and class-specific distance measures}

Given a training data set of time series, we can learn a single Mahalanobis distance measure from all time series, irrespective of their class labels (henceforth global Mahalanobis  distance measure). We have a single matrix $M$. In this case the 1-NN classification algorithm works as follow: given a candidate time series $\mathbf{x}$ we seek $\mathbf{y}$ from the training set such that $D_M(\mathbf{x},\mathbf{y})$ is minimal. We then classify $\mathbf{x}$ in the same class as $\mathbf{y}$.  

Otherwise, we may learn one Mahalanobis distance measure per class of time series. In this class-specific approach, the covariance matrix is computed solely from the time series of one class. Hence, for each class, we get one distance  measure: class $i$ gets distance measure $D_{M_i}$. We have one matrix ($M_i$) per class ($i$).
Given $\mathbf{y}$ from the training set, let $c(\mathbf{y})$ be the class of $\mathbf{y}$.
Given a candidate time series $\mathbf{x}$,
we classify it by finding
$\mathbf{y}$  such that its distance  to $\mathbf{x}$, $D_{M_{c(\mathbf{y})}}(\mathbf{x},\mathbf{y})$, is minimal. Hence, we not only compare the candidate time series $ \mathbf{x}$ with time series from different classes, but we also use different distance measures.

Thus, finally, we consider six types of Mahalanobis distance measures for 1-NN classification: two localities (global or class-specific) and three estimators (pseudoinverse, shrinkage, or diagonal).

\section{Experiments}
\label{sec:experiments}

The main goal of our experiments is to evaluate Mahalanobis distance measures and the class-specific approach on time series. More specifically, we ask the following questions. 
\begin{itemize}
\item Of all the possible applications of the Mahalanobis distance measures (pseudoinverses, shrinkage or diagonal; class-specific or global), which one offers the best 1-NN classification accuracy? (\S~\ref{sec:bestfor1nn})
\item How do Mahalanobis distance measures compare with state-of-the-art alternatives such as DTW or LMNN? 
(\S~\ref{sec:compcompetitive})
\item One of the simplest and most common distance measures, the Euclidean distance,  is sometimes difficult to surpass for 1-NN classification of time series. To assess this effect,  we ask how the relative accuracy  of the Mahalanobis distance measure changes as we increase  the number of instances per class in the training set. (\S~\ref{sec:numberofinstances})
\end{itemize}  

We begin all tests with a training data set comprising several classes of time series. When applicable,
distance measures are learned from this data set. We then attempt to classify some test
data using 1-NN\@.
We define the  \emph{classification error} to be the percentage of misclassified instances whereas the 
\emph{accuracy} is the percentage of properly classified instances.

The code for the experiments is available online~\citep{ourmatlabcode} with instructions on how the results can
be reproduced. For LMNN, we use the source code provided by \cite{lmnncode} for the experiments
with default parameters.
For the DTW, we find the best monotonic  matching $\Gamma$ minimizing 
${\sum_{(i,j)\in \Gamma} |x_i - y_j|^2}$.
The computational cost of the DTW is sometimes a challenge~\citep{Salvador2007561}.
To alleviate this problem, several strategies have been proposed
including lower bounds and R$^{\star}$-tree indexes~\citep{ratanamahatana2005tmd,Lemire:2009:FRT:1542560.1542887,Ouyang:2010:HDS:1884499.1884520}.
For our purposes, we use a quadratic-time dynamic programming algorithm. In contrast, the Euclidean and diagonal Mahalanobis distance measures only require linear time.
We ran the experiments  on a MacBook~Pro laptop with a 2.3~GHz Intel Core~i5 processor and 8~GB of RAM\@. All code ran on  Matlab~R2011a.

\subsection{Data sets}
We use the UCR time series classification benchmark~\citep{keoghbenchmark} for our experiments as it includes
diverse time series data sets from many domains. 
It has predefined training-test splits for the experiments (see Table~\ref{tab:boring}),
so the results can be compared across different papers. 
Most of the data sets are z-normalized: that is, the time series have zero mean and a variance of one.
We removed the two data that are not z-normalized by default (Beef and Coffee).
Indeed, z-normalization improves substantially the classification accuracy~---~irrespective of the chosen distance measure.
Thus, for fair results, we should z-normalize them, but this may create confusion with previously
reported numbers. 
We also removed the Wafer data set as all distance measures classify it nearly perfectly.
The remaining 17~data sets were used for the comparison of different methods.

\begin{table}
\caption{Number of classes, number of instances in both the training and testing sets, and the length of the time series in each data set. \label{tab:boring}}
\centering
\begin{tabular}{l|cccc}\hline
Data set           & classes  & training set & testing set & length ($n$) \\\hline
50 words &	 50 &	 450 	& 455 &	 270\\
 Adiac  & 37 &	390 &	391 &	176\\
CBF  &     3 	& 30 &	 900 &	 128 \\
ECG  &    2 & 	100 & 	100 & 	96\\
Fish&  7 & 	175 & 	175 & 	463\\
Face (all) &  14 	&  560 & 	 1\,690 & 	 131\\
Face (four) &  4 	&  24 & 	 88 	&  350\\
Gun-Point  &  2 & 	 50 & 	 150 & 	 150\\
Lighting-2 &   2 	&  60 & 	 61 	&  637\\
Lighting-7 &   7 & 	 70 & 	 73 & 	 319\\
OSU Leaf  &  6 & 	 200 & 	 242 & 	 427\\
OliveOil & 4& 30& 	30& 	570\\
Swedish Leaf &  15&  	 500&  	 625&  	 128\\
Trace  & 4 & 	 100 & 	 100 & 	 275\\
Two Patterns  & 4&  	 1\,000 & 	 4\,000&  	 128\\
Synthetic Control&  6 	& 300 & 300 & 	60\\
Yoga       &       2 & 	300 	& 3\,000 & 	426\\\hline
\end{tabular}
\end{table}

\subsection{Best Mahalanobis distance measure for 1-NN accuracy}
\label{sec:bestfor1nn}

We compare the various Mahalanobis distance measures in Table~\ref{tab:ourresultsforMahalanobis}. We have left out the Moore-Penrose pseudoinverse, because its error rates were twice as high on average compared to the other variants.
What is immediately apparent is that the class-specific measures give better classification results.

The diagonal Mahalanobis has a smaller classification error and is considerably faster (3.7~min compared to 5.5~min on the whole data set), but the shrinkage estimate yields significantly better results for several data sets (e.g.~Adiac and Fish).
Thus, out of the six variations, we recommend the class-specific covariance-shrinkage estimate and the class-specific diagonal Mahalanobis distance measures. 

\begin{table}
\caption{\label{tab:ourresultsforMahalanobis}Classification error for the various Mahalanobis distance measures.}
\centering
\begin{tabular}{l|cccc}\hline
\multirow{2}{*}{Data set}           & \multicolumn{2}{c}{Shrinkage}  & \multicolumn{2}{c}{Diagonal}              \\
             & global &  class-specific         & global & class-specific \\\hline
50 words			& 0.34		& 0.71	&  0.34			&  \textbf{0.32}\\
Adiac               & 0.33     & \textbf{0.30}          &  0.37         &  0.36\\
CBF                 &  0.34     & 0.06 &  0.16         &   \textbf{0.05} \\
ECG                 &  0.12     & 0.10          &0.10           & \textbf{0.08}     \\
Fish                &  0.31     & \textbf{0.15}          & 0.19 & 0.18  \\
Face (all)          &  0.32     & 0.27	&  0.32         & \textbf{0.25}\\ 
Face (four)         &  0.27     & \textbf{0.16}          & \textbf{0.16}         &0.17  \\ 
Gun-Point           &  0.12     & 0.14          & \textbf{0.10}          & 0.11 \\
Lighting-2          &  0.31     & 0.30 &  \textbf{0.25}          &  \textbf{0.25}  \\
Lighting-7          &  0.55     & 0.32          & 0.36          &  \textbf{0.23} \\
OSU Leaf            &  0.68     & 0.69 & \textbf{0.46}          & \textbf{0.46}    \\
OliveOil          &0.17			& 0.20  & 0.17          &\textbf{0.13}  \\
Swedish Leaf        & 0.24     & \textbf{0.15}          & 0.21          & 0.18 \\
Trace               &  0.40     & 0.12 & 0.21          & \textbf{0.07}   \\
Two Patterns        &  0.12     & 0.12 & 0.12          &  0.12\\
Synthetic Control   &  0.23     & \textbf{0.08} & 0.13          &  0.09\\
Yoga              & 0.26 		& 0.22	& \textbf{0.17}         & \textbf{0.17}	 \\\hline
\# of best errors   & 0 & 5 & 5 & 10 \\ \hline
\end{tabular}
\end{table}

\subsection{Comparing competitive distance measures}
\label{sec:compcompetitive}

How do the class-specific Mahalanobis distance measures behave in comparison with  competitive distance measures? Computationally,
the diagonal Mahalanobis is inexpensive compared to schemes such as the DTW or LMNN\@. 
Regarding the 1-NN classification error rate, we give the results in Table~\ref{tab:ourresults}. 
As expected~\citep{1454226}, no distance measure is better on all data sets. However, because the diagonal Mahalanobis distance measure is closely related
to the Euclidean distance, we compare their classification accuracy.
In two data sets, the Euclidean distance outperformed the class-specific Mahalanobis distance measures and only by small differences (0.09 versus 0.10--0.12). Meanwhile, the class-specific diagonal Mahalanobis distance measures outperformed the Euclidean distance 12~times, and sometimes by large margins (0.07 versus 0.24 and 0.05 versus 0.15).
The LMNN is also competitive: its classification error is sometimes half that of the Euclidean distance.

The DTW has the lowest error rates and provides best results for half of the data sets, but it is much slower than Mahalanobis distance measures.
It takes 3.7~min (diagonal) and 5.5~min 
 (covariance shrinkage) to compute the Mahalanobis results on the whole data set.
As expected, the diagonal Mahalanobis is nearly as fast as the Euclidean distance (3.5~min).
The LMNN takes 18~min and the DTW runs for 18~hours.
The DTW is at least two orders of magnitude slower than the diagonal Mahalanobis on all 17~data sets.  


\begin{table}
\caption{\label{tab:ourresults}Classification errors for some competitive schemes. For all distance measures, we use 1-NN classification. For the 50~words data set, the LMNN computation fails because it has a class with only one instance. For this table, we used the class-specific Mahalanobis distance measure.}
\centering 
\begin{tabular}{l|cccccc}\hline
\multirow{1}{*}{Data set}           & \multirow{1}{*}{Euclidean} & \multirow{1}{*}{DTW}           & \multicolumn{2}{c}{Mahalanobis} & \multicolumn{1}{c}{LMNN}  &  \\
             &  &     &   shrink.   & diag.  &     \\\hline
50 words			&  0.37		& \textbf{0.31}	& 0.71	&   0.32				& --			 \\
Adiac               &  0.39     & 0.40          & 0.30         &   0.36               & \textbf{0.23}     \\
CBF                 &  0.15     & \textbf{0.00} & 0.06         &   0.05               &  0.15           \\
ECG                 &  0.12     & 0.23          & 0.10           & \textbf{0.08}        	&  0.10          \\
Fish                &  0.22     & 0.17      	& 0.15	     & 0.18                 &  \textbf{0.13} \\
Face (all)          &  0.29     & 0.19			& 0.27        & 0.25         &  \textbf{0.16}         \\ 
Face (four)         &  0.22     & 0.17          & \textbf{0.16}     & 0.17                  & \textbf{0.16}  \\ 
Gun-Point           &  0.09     & 0.09          & 0.14        & 0.11                 &\textbf{0.05}      \\
Lighting-2          &  0.25     & \textbf{0.13} & 0.30          & 0.25                 &  0.41           \\
Lighting-7          &  0.42     & 0.27          & 0.32         &  \textbf{0.23}       &  0.51           \\
OSU Leaf            &  0.48     & \textbf{0.41} & 0.69         & 0.46                 & 0.57            \\
OliveOil          &\textbf{0.13}&\textbf{0.13}  & 0.20         &\textbf{0.13}         & \textbf{0.13}    \\
Swedish Leaf        &  0.21     & 0.21          & \textbf{0.15}   & 0.18        & 0.21       \\
Trace               &  0.24     & \textbf{0.00} & 0.12          & 0.07                 & 0.20           \\
Two Patterns        &  0.09     & \textbf{0.00} & 0.12          &  0.12                & 0.05		  	\\
Synthetic Control   &  0.12     & \textbf{0.01} & 0.08          & 0.09                 & 0.03         \\
Yoga              & 0.17 		& \textbf{0.16}	& 0.22          & 0.17					& 0.18		 \\
\hline
\# of best errors   & 1 & 9 & 2 & 3 & 6 \\ \hline
\end{tabular}
\end{table}

\subsection{Effect of the number of instances per  class}
\label{sec:numberofinstances}
Whereas Table~\ref{tab:ourresults} shows that the Mahalanobis distance measures are far superior to the Euclidean distance on some data sets, this result
 is linked to the number of instances
per class. For example, on the Wafer data set (which we removed), there are many instances per class (500), and correspondingly, all distance measures give a negligible classification error.

Thus, we considered three different synthetic time-series data-set generators with varying numbers of instances per class:
\begin{itemize}
\item Cylinder-Bell-Funnel (CBF)~\citep{921732},
\item   Control Charts (CC)~\citep{pham1998ccp} and
\item  Waveform~\citep{breiman1998car}.
\end{itemize}
 The CC data is made of 6~classes containing time series made of 60~data points; CBF is made of 3~classes and its time series have 128~data points; Waveform has 3~classes and its time series are made of 21~data points.  All time series are z-normalized (zero mean and a variance of one).
The CBF data set from Tables~\ref{tab:ourresultsforMahalanobis}
 and~\ref{tab:ourresults} was generated from the same data-set model, except that  we vary the number of time series (see Appendix~\ref{appendix:cbf}).

\begin{figure}
\centering\includegraphics[width=0.8\columnwidth]{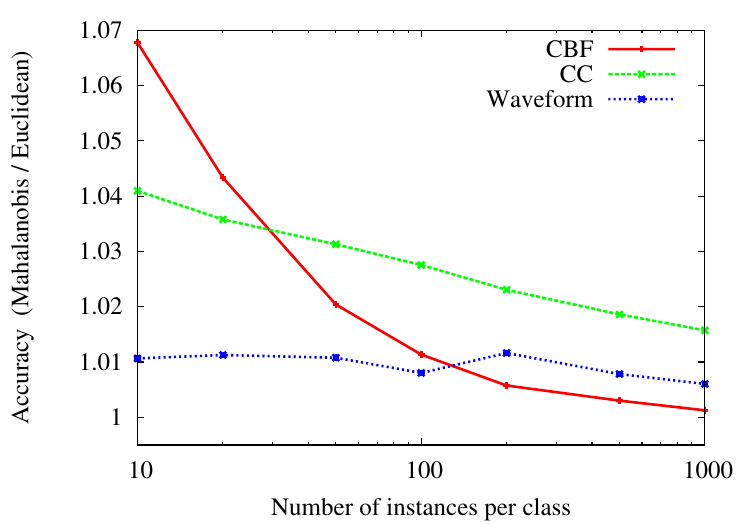}
\caption{\label{fig:synthdatasets}Ratios of the 1-NN classification accuracies using the class-specific diagonal Mahalanobis and Euclidean distance measures}
\end{figure}

Test sets have 1\,000~instances per class whereas training sets have between 10 to 1\,000 instances.
We repeated each test ten times, with different training sets.
Fig.~\ref{fig:synthdatasets} shows that whereas the class-specific diagonal Mahalanobis distance measures
are superior to the Euclidean distance when there are few instances, this benefit
is less significant as the number of instances increases. Indeed, the classification
accuracy of the Euclidean distance grows closer to perfection and it becomes more difficult
for alternatives to be far superior.

%
%

\section{Conclusion}
The Mahalanobis distance measures have received little attention for time series classification and we are not
surprised given their poor performance
as a 1-NN classifier when used in a straight-forward manner. However, by learning one Mahalanobis distance measure per class we get a competitive classifier when using either covariance shrinkage or a diagonal approach.
Moreover, the diagonal Mahalanobis distance measure is particularly appealing computationally: we only
need to compute the variances of the components.
Meanwhile, we get good results with the LMNN on time series data, though it is more expensive. 
The DTW is superior, but it is one to two orders of magnitude slower.

\begin{acknowledgements}
This work is supported by NSERC grant 261437.
\end{acknowledgements}

\bibliographystyle{spbasic}
\bibliography{Mahalanobisclassification}

\appendix

\section{Choice of the parameter $\lambda$ in covariance shrinkage}
\label{sec:choiceoflambda}
Covariance shrinkage (see \S~\ref{sec:computing}) requires
the choice of a parameter $\lambda\in (0,1]$, which should
be sufficiently large so that $\lambda T + (1- \lambda) C$
is numerically invertible. We choose $T$ to be the diagonal of the 
covariance matrix $C$ ($T=\mathrm{diag}(C)$). 

To give the formula for $\lambda$ proposed by \cite{schafer2005shrinkage}, we need to introduce some technical 
notation.
Given a family of time series
$\mathbf{x}^{(1)}, \mathbf{x}^{(2)},\ldots, \mathbf{x}^{(N)}$,
we write the average of the $i^{\mathrm{th}}$ component
as $\bar z_i= \sum_{k=1}^N \mathbf{x}^{(k)}_i$. We write
\begin{eqnarray*}
w_{kij}= (\mathbf{x}^{(k)}_i-\bar z_i)(\mathbf{x}^{(k)}_j-\bar z_j)\end{eqnarray*}
and $w_{ij}=\frac{1}{N}\sum_{k=1}^N w_{kij}$.
Moreover, we write 
\begin{eqnarray*}\widehat{\mathrm{Var}}(c_{ij})=\frac{N}{(N-1)^3}\sum_{k=1}^N (w_{kij}-w_{ij})^2.
\end{eqnarray*} Finally, we have
\begin{eqnarray*}
\lambda^{\star}=\frac{\sum_{i\neq j} \widehat{\mathrm{Var}}(c_{ij})}{\sum_{i\neq j} c_{ij}^2}
\end{eqnarray*}
where $c_{ij}$ are the components of the (sample) covariance matrix $C$ (see \S~\ref{sec:mathstuff}).
We set $\lambda$ to  $\lambda^{\star}$ when  $\lambda^{\star}\leq 1$.
Otherwise, we set $\lambda=1$.

We experimented informally with different values of $\lambda$ (using the
data sets from from \S~\ref{sec:experiments})  and found that the choice
preconised by \cite{schafer2005shrinkage} was reasonable. That is, we did not find a case where a different value of  $\lambda$ gave much better classification accuracy.

\section{The Cylinder-Bell-Funnel (CBF) data model}
\label{appendix:cbf}

Consider the original CBF data model~\citep{921732}. We can use it to generate time series of three possible classes.
In the case where we have only 10~time series for each class in the training data set and a large number of time series in the test set (1000), we find that, over ten tests, the average 1-NN classification error rate  is 0.20 ($\sigma = 0.04$) for the Euclidean distance and  0.15 ($\sigma = 0.03$) for the diagonal Mahalanobis distance measure. These results are difficult to reconcile with Table~\ref{tab:ourresults} where we used a similar number of CBF time series provided by \cite{keoghbenchmark} and where we report error rates of 0.15 and 0.05. Indeed, the difference in error rate for the diagonal Mahalanobis distance measure exceeds 3~standard deviations.

After inspection, we found that the CBF data model used by \cite{keoghbenchmark} differs from the original presentation by \cite{921732}.
 They both generate time series using random functions of the form: 
 $c(i)= (6 + \eta) \cdot \chi_{[a,b]}(i)+\varepsilon(i)$,
$b(i)= (6 + \eta) \cdot \chi_{[a,b]}(i) \cdot (i-a)/(b-a)+\varepsilon(i)$ and 
$f(i)= (6 + \eta) \cdot \chi_{[a,b]}(i) \cdot (b-i)/(b-a)+\varepsilon(i)$ where $i =1,\ldots,128$ and $\chi_{[a,b]}$ is the characteristic function. They both use standard normal variates for $\eta$ and $\varepsilon(i)$, and uniformly distributed $a$ integer values in $[16,32]$. However, whereas  \cite{921732} states that $b-a$ obeys an integer-valued uniform distribution on $[32,96]$, we found that \cite{keoghbenchmark} generated their CBF data so that $b-32$ is an integer-valued uniform distribution on $[32,96]$.

If we  adopt the \cite{keoghbenchmark} variation, the classification error rates go down: 0.16 ($\sigma = 0.03$) for the  Euclidean distance and 0.10 ($\sigma = 0.04$) for the diagonal Mahalanobis distance measure. These results are nearly within a standard deviation of the results presented in  Table~\ref{tab:ourresults} for CBF.

\end{document}